\documentclass[pdflatex,sn-mathphys-num]{sn-jnl}

\usepackage{graphicx}%
\usepackage{multirow}%
\usepackage{amsmath,amssymb,amsfonts}%
\usepackage{amsthm}%
\usepackage{mathrsfs}%
\usepackage[title]{appendix}%
\usepackage{xcolor}%
\usepackage{textcomp}%
\usepackage{manyfoot}%
\usepackage{booktabs}%
\usepackage{algorithm}%
\usepackage{algorithmicx}%
\usepackage{algpseudocode}%
\usepackage{listings}%
\usepackage{tabularx} %
\usepackage{makecell} %
\usepackage{multirow}
\usepackage{booktabs}
\usepackage{tabularx}
\usepackage{array}
\usepackage{tabularx}
\usepackage{booktabs}
\usepackage{multirow}
\usepackage{makecell}
\usepackage{caption}
\usepackage{array}    
\usepackage{tabularx}
\usepackage{booktabs}
\usepackage{array,tabularx,booktabs,multirow,makecell,xcolor}

\newcolumntype{Y}{>{\centering\arraybackslash}X} 
\newcolumntype{L}{>{\raggedright\arraybackslash}X} 

\theoremstyle{thmstyleone}%
\theoremstyle{thmstyletwo}%

\theoremstyle{thmstylethree}%

\begin{document}

\title[Article Title]{RiskNet: A large-scale dataset of AI risk incidents from news with alignment and multi-dimensional annotations}


\author*[1,2]{\fnm{Leihan} \sur{Zhang}}\email{zhangleihan@gmail.com}

\author[1]{\fnm{Wecheng} \sur{Ye}}\email{weichengye@bupt.edu.cn}

\author[1]{\fnm{Xianlong} \sur{Ma}}\email{maxianlong@bupt.edu.cn}

\author[1]{\fnm{Haochuan} \sur{Liu}}\email{liuhaochuan@bupt.edu.cn}

\author[1]{\fnm{Yang} \sur{Li}}\email{leonli@bupt.edu.cn}

\author[1]{\fnm{Qianyu} \sur{Zhang}}\email{zhangqyn@bupt.edu.cn}
\author[1]{\fnm{Jinliang} \sur{Chen}}\email{434908141@qq.com}
\author*[1,2]{\fnm{Qiang} \sur{Yan}}\email{yan@bupt.edu.cn}

\affil[1]{\orgdiv{School of Economics and Management}, \orgname{Beijing University of Posts and Telecommunications}, \orgaddress{\street{10 Xitucheng Road, Haidian District}, \city{Beijing}, \postcode{100876‌},  \country{China}}}

\affil[2]{\orgdiv{Beijing Key  Laboratory  of Multimodal Data  Intelligent Perception  and Governance}, \orgaddress{\street{10 Xitucheng Road, Haidian District}, \city{Beijing}, \postcode{100876‌},  \country{China}}}



\abstract{
As artificial intelligence (AI) systems are increasingly deployed across socially consequential domains, reports of AI-related harms and failures have grown in frequency and diversity. Although existing governance frameworks articulate high-level principles for responsible AI, large-scale empirical resources for tracking and analyzing real-world AI risk incidents remain limited. Existing incident collections are often manually curated, relatively small in scale, and insufficient for continuous, data-driven monitoring and downstream computational analysis.

To address this need, we present RiskNet, a large-scale dataset of AI risk incidents constructed from large-scale multilingual news sources. RiskNet applies a structured pipeline for AI risk news identification, event-level report screening, incident alignment, and multi-dimensional incident classification. The resulting resource organizes dispersed news reports into incident-centered records and provides benchmark datasets for event classification, incident alignment, and incident-level risk labeling. In its current release, RiskNet covers hundreds of millions of source records and yields a large-scale collection of AI risk-related reports, including aligned incident clusters and annotated benchmark subsets. The dataset is also accessible through an online platform for browsing and exploration.

We describe the data sources, processing workflow, taxonomy design, and technical validation of the resource. RiskNet is intended to support downstream research on AI safety, governance, risk analysis, and benchmarking, as well as longitudinal and cross-source analyses of AI-related harms. By providing a structured and reusable empirical resource, RiskNet helps bridge the gap between high-level governance principles and the documented realities of AI risk incidents.
}
\keywords{AI risk, Dataset, Classification, Entity alignment, Ontology}



\maketitle 

\section{Background \& Summary}\label{sec1}

Artificial intelligence (AI) is now used in many high-impact settings, including healthcare, education, finance, transportation, and public governance. As these systems become more capable and more widely deployed, concerns about their risks have also grown. Recent assessments show that even highly capable general-purpose AI (GPAI) systems can behave unpredictably and introduce new forms of systemic risk~\cite{bengio_safety_2026}. Reported incidents already include algorithmic bias, misinformation, privacy leakage, safety failures, and other social harms~\cite{bengio_managing_2024,winfield2018ethical,Brent2026the}. Evidence from 2025 further suggests that AI-related incidents are continuing to increase, driven in part by deepfake-enabled fraud and the wider use of autonomous agentic systems~\cite{time_ai_harm_2025,staufer_agent_2026}. These trends make it increasingly important to build structured empirical resources for tracking and analyzing real-world AI risk incidents.

Governments and regulatory bodies have responded by introducing high-level frameworks for AI governance, including the U.S. \textit{AI RMF 1.0} \cite{tabassi_artificial_2023}, the EU \textit{AI Act} \cite{euAIAct2024}, and China’s \textit{GenAI Measures} \cite{chinaGenAI2023}. These frameworks define important principles such as safety, fairness, transparency, and accountability. However, applying these principles to real incidents remains difficult. Evidence is scattered across multilingual news reports, described in inconsistent ways, and often spread across multiple reports with different actors, causes, and outcomes \cite{golpayegani_be_2023,yu_understanding_2025}. The problem is made harder by a clear language gap in AI safety evaluation and moderation, since many existing tools work much better in English than in non-English or low-resource language settings~\cite{yong_multilingual_2025}. As a result, AI risk incidents are still difficult to collect, compare, and study systematically \cite{bengio_managing_2024,de_miguel_velazquez_decoding_2024}.

Previous work has provided useful ways to organize AI risk information. Some studies define taxonomies from regulatory, ethical, or application perspectives, such as the U.S. AI RMF and the EU AI Act \cite{tabassi_artificial_2023,euAIAct2024}. Others try to make these frameworks more operational for empirical analysis, including AIR-Bench 2024 \cite{zeng_airbench_2024}, the AI Risk Repository \cite{slattery_ai_2024,mylius_repository_2025}, and empirical taxonomies for generative-AI harms in specific populations \cite{yu_understanding_2025}. These efforts help describe the AI risk landscape, but they do not mainly function as large-scale, multilingual, incident-centered datasets that can align multiple reports about the same event and support downstream computational analysis.

Several public resources have also been developed to document AI-related incidents, including AIID \cite{mcgregor_preventing_2020}, AIAAIC \cite{aiaaic2023}, the AI Risk Repository \cite{slattery_ai_2024}, and broader reviews of incident repositories and case collections \cite{knight_learning_nodate,turri_why_2023}. These resources have supported case discovery, qualitative analysis, and regulatory reflection on AI harms \cite{lupo_risky_2023}. Domain-specific studies further suggest that important safety and ethical issues may remain underrepresented in general-purpose repositories \cite{gipiskis_health_2026}. At the same time, recent analyses of incident databasing have identified persistent problems, including uneven source coverage, ambiguous incident boundaries, uncertainty in reported facts, and continued dependence on manual curation \cite{paeth_lessons_2024}. New work is beginning to address these limitations by improving incident reporting standards, automating report linking and extraction, and supporting more structured incident analysis \cite{rao_riskrag_2025,oecd_reporting_2025,bieringer_taxonomy_2025,russo_automating_2025,richards_incidents_2025,popchanovska_mitigation_2026,chu_llm_incident_2026}. Even so, there remains a need for a resource that combines broad coverage, multilingual collection, incident-level alignment, and reusable benchmark annotations.

To address this need, we present \textbf{RiskNet}, an open dataset of AI risk incidents. RiskNet collects multilingual news reports on AI-related harms and organizes them through AI risk news identification, event-level report screening, incident alignment, and multi-dimensional incident classification. The resulting resource is designed to support research in AI safety, governance, social science, and information science. In particular, it supports cross-lingual and longitudinal analysis of AI risk incidents, benchmark construction for incident alignment and classification, and structured study of risk domains and causal patterns.

The main components of RiskNet include:
\begin{itemize}
  \item a large-scale collection of multilingual, multi-source AI risk incident reports with incident-level alignment;
  \item a unified risk taxonomy and benchmark dataset supporting multi-dimensional labels such as domain, cause, and severity; and
  \item an openly accessible platform intended to support data exploration and downstream reuse.
\end{itemize}

\section{Methods}\label{sec2}

\subsection{Overview}\label{subsec2_overview}

RiskNet is constructed from large-scale, heterogeneous news sources through three main stages: 
(1) AI risk incident identification, 
(2) incident alignment, and 
(3) multi-dimensional incident classification. 
The resulting resource organizes dispersed AI risk reports into an incident-centered dataset 
suitable for downstream reuse in AI safety, governance, and social-scientific research.

We formalize these three stages as follows. 
Let \(a\) denote a news article, and let \(\mathcal{A}\) represent the corpus of collected raw news records.

\textbf{(1) AI Risk Incident Identification.} 
We define a filtering function 
\[
g: \mathcal{A} \rightarrow \{0,1\},
\]
which determines whether an article describes an AI risk incident. 
The resulting subset of incident-related articles is:
\[
\mathcal{A}_{incident} = \{ a \in \mathcal{A} \mid g(a) = 1 \}.
\]

\textbf{(2) Incident Alignment.} 
We construct a set of real-world incidents:
\[
\mathcal{I} = \{ inc_1, inc_2, \dots, inc_N \},
\]
where each \(inc_i\) represents a distinct incident aggregated from multiple reports. 
Each article in \(\mathcal{A}_{incident}\) is assigned to exactly one incident via an alignment function
\[
f: \mathcal{A}_{incident} \rightarrow \mathcal{I},
\]
where multiple articles may correspond to the same incident.

\textbf{(3) Multi-dimensional Incident Classification.} 
We further assign labels to each incident along multiple dimensions. 
Let \(\mathcal{C}^k\) denote the label space for dimension \(k\), where
\[
k \in \{\textit{entity}, \textit{intent}, \textit{time}, \textit{risk level}, \textit{domain}, \textit{subdomain}\}.
\]
We define a classification function
\[
h_k: \mathcal{I} \rightarrow \mathcal{C}^k,
\]
which assigns each incident to a class in dimension \(k\).

This formulation provides a unified view of the pipeline from article-level observations to structured, incident-level representations with multi-dimensional annotations.

\subsection{Data Sources}\label{subsec2_sources}

RiskNet integrates AI risk-related content from multiple source types to improve temporal coverage, linguistic diversity, and source complementarity. 
The data sources include publicly available AI incident datasets, open and collected news corpora, and a licensed commercial news dataset.

\begin{itemize}
    \item \textbf{Publicly available AI incident datasets curated by research organizations ($P$):}
    \begin{itemize}
        \item \textbf{AI and Algorithmic Incident and Controversies (AIAAIC)} curates over 4,500 reports~\cite{aiaaic2023}.
        \item \textbf{AI Incident Database (AIID)} collects approximately 5,000 reports through community contributions and web-based collection~\cite{slattery_ai_2024}.
    \end{itemize}

    \item \textbf{Open-source and collected news corpora:}
    \begin{itemize}
        \item \textbf{CommonCrawl News ($C$)} (2022--2025): 674M news articles, subsequently processed through AI relevance filtering and AI risk incident identification~\cite{commoncrawlnews}.
        \item \textbf{OpenNewsArchive ($O$)} (2023): 8.8M Chinese-language news articles~\cite{he2024opendatalabempoweringgeneralartificial}.
        \item \textbf{Hot-list Rankings ($H$)} (2021--2025): 340K trending topics collected from platforms such as Douyin, Weibo, Zhihu, Toutiao, and v2ex~\cite{hot-hub}.
        \item \textbf{News websites ($N$)} (2024): approximately 800 articles retrieved through keyword-based crawling from news portals such as Baidu News.
    \end{itemize}

    \item \textbf{Licensed commercial news dataset ($L$):}
    \begin{itemize}
        \item \textbf{China News (commercial procurement)} (2023.10--2025.12): 2.98M AI-related negative news articles, further screened for AI risk relevance.
    \end{itemize}
\end{itemize}

The reported time ranges reflect the current snapshot of the collected datasets. 
RiskNet is an evolving resource that will be continuously updated and maintained to incorporate newly available data.

\subsection{AI Risk Incident Identification}\label{subsec2_identification}

\begin{figure}
    \centering
    \includegraphics[width=\linewidth]{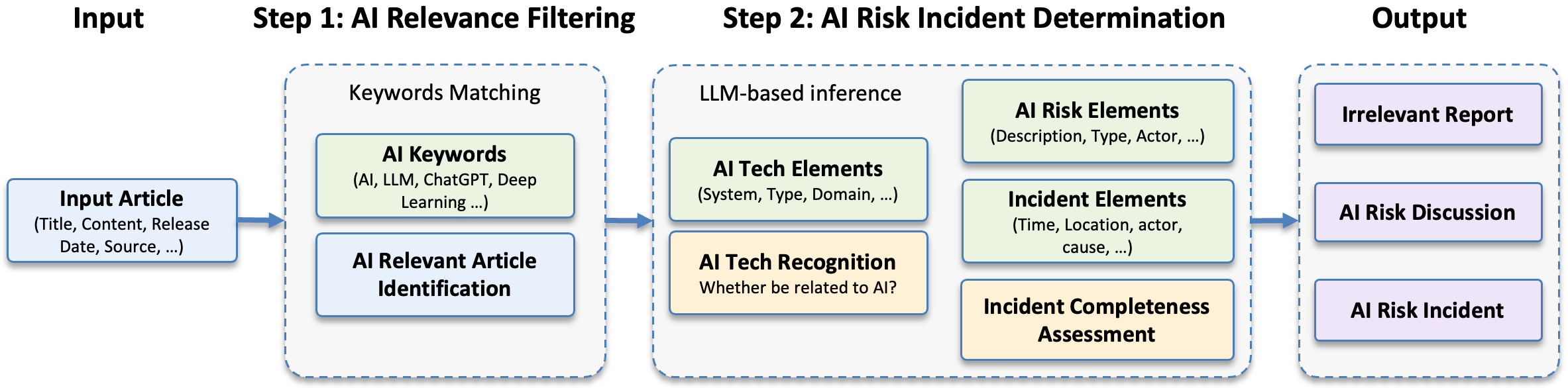}
    \caption{The LLM-based AI Risk Incident Identification Pipeline.}
    \label{fig:Identification}
\end{figure}

As shown in Fig.~\ref{fig:Identification}, AI risk incident identification is conducted in two stages: 
(1) AI relevance filtering and 
(2) AI risk incident determination.

\subsubsection{Step 1: AI Relevance Filtering}

\begin{algorithm}[t]
\caption{AI Relevance Filtering and AI Risk Incident Identification}
\label{alg:identification}
\begin{algorithmic}[1]
\Require Source collections \(H, C, O, N, P, L\), AI keyword lexicon \(K\), top-\(K\) keyword extractor
\Ensure AI risk incident article set \(\mathcal{A}_{incident}\)

\State Initialize \(\mathcal{A}_{incident} \leftarrow \emptyset\)
\For {$X \in \{H, C, O, N, P, L\}$}
    \If{$X = H$}
        \State Extract top-\(K\) keywords from each hot topic title
        \State Retain topics whose extracted keywords overlap with \(K\)
        \State Retrieve related news reports \(X_{AI}\) through a search engine
    \ElsIf{$X = N$}
        \State Query news portals using the AI keyword lexicon \(K\) to obtain \(X_{AI}\)
    \ElsIf{$X = P$ \textbf{or} $X = L$}
        \State Use the source collection directly as candidate set \(X_{AI}\)
    \Else
        \State Extract top-\(K\) keywords from each candidate text
        \State Retain articles with at least one keyword matched in \(K\), denoted as \(X_{AI}\)
    \EndIf
    \State Apply LLM-based incident determination to \(X_{AI}\)
    \State Retain articles identified as AI risk incidents and update
    \State \hspace{1em} \(\mathcal{A}_{incident} \leftarrow \mathcal{A}_{incident} \cup \text{LLM}(X_{AI})\)
\EndFor
\State \Return \(\mathcal{A}_{incident}\)
\end{algorithmic}
\end{algorithm}

Algorithm~\ref{alg:identification} summarizes the unified pipeline for AI relevance filtering and AI risk incident identification.

Step 1 uses a maintained AI keyword lexicon containing more than 100 terms related to AI technologies, systems, models, and products. 
For each candidate text, we extract its top-\(K\) keywords and match them against the lexicon. 
A text is considered AI-related if at least one extracted keyword matches an entry in the lexicon. 
This stage is designed as a high-recall filtering step to reduce the search space before more fine-grained LLM-based screening.

Publicly available incident datasets (AIID and AIAAIC) were incorporated directly as source collections, while the remaining sources were processed through the same relevance-filtering principle. 
For large-scale open news corpora, keyword-based filtering was first applied to identify AI-related candidates. 
These candidate reports were then passed to a large language model (LLM) for AI risk incident determination. 
For downstream event-oriented analysis, the same LLM workflow also extracted core incident elements, which were subsequently used to distinguish incident-level reports from risk-related discussions.

The implementation varies slightly across source types while preserving the same filtering criterion:

\begin{enumerate}
    \item \textbf{CommonCrawl News and OpenNewsArchive.}  
    For article collections with title and body text, we extracted top-\(K\) keywords from each text and matched them against the AI keyword lexicon. Articles with at least one matched keyword were retained as AI-related candidates and then evaluated by the LLM.

    \item \textbf{Hot-list Rankings.}
    Since this source contains only topic titles rather than full reports, we applied the same keyword matching procedure to topic titles. Topics identified as AI-related were then expanded into associated news reports through a search engine, and the retrieved reports were further screened by the LLM.

    \item \textbf{News websites.}
    For news portals, the maintained AI keyword lexicon was directly used as the query set for site-level retrieval. The retrieved reports were then passed to the same LLM-based incident identification procedure.

    \item \textbf{Public AI incident datasets and licensed commercial data.}
    These sources were directly incorporated as candidate collections because they had already been pre-curated as AI-related or risk-relevant sources. We further applied the same LLM-based screening procedure to normalize inclusion criteria and identify records suitable for RiskNet.
\end{enumerate}

\subsubsection{Step 2: AI Risk Incident Determination}

The second stage determines whether an AI-related article should be retained as an event-level AI risk incident or treated as a discussion-level AI risk report. 
This distinction is motivated by the requirements of reliable incident alignment.

Event-level AI risk reports typically contain sufficiently grounded and structured event information, including identifiable actors, AI systems, spatio-temporal context, and observable outcomes. 
Such reports provide the necessary anchors for accurate cross-document aggregation and high-precision incident alignment.

In contrast, many AI risk-related articles consist of policy interpretation, expert commentary, ethical debate, or general discussion. 
Although these discussion-level reports are valuable for understanding public discourse and broader societal perspectives, they often lack concrete and localized event details. 
As a result, incorporating them into the alignment process would introduce substantial ambiguity while providing limited benefits for incident reconstruction.

Therefore, RiskNet prioritizes event-level reports for incident alignment, while treating discussion-level reports as auxiliary data for complementary analysis rather than core alignment inputs.

Accordingly, after Step 1 identifies AI-related candidates, we apply an LLM to determine whether the article explicitly describes an AI risk and, if so, whether it contains sufficiently grounded incident information. 
This stage therefore serves two purposes simultaneously: 
(1) identifying AI risk relevance, and 
(2) distinguishing event-level incident reports from discussion-level risk discussions.

\begin{table*}[t]
\centering
\small
\setlength{\tabcolsep}{8pt}
\renewcommand{\arraystretch}{1.5}
\caption{Structured extraction schema for AI risk incident identification and representation}
\label{tab:extraction_schema}
\begin{tabularx}{\textwidth}{@{} >{\bfseries}m{3.2cm} >{\raggedright\arraybackslash}X @{}}
\toprule
\textbf{Dimension} & \textbf{Extracted Attributes} \\
\midrule
AI System \& Application & 
AI systems (ai\_system\_list), 
system type (ai\_system\_type\_list), 
application domain (ai\_system\_domain\_list) \\

Event Context & 
Time (event\_time\_start, event\_time\_end), 
location (event\_country, event\_province, event\_city), 
actors (actor\_main, actor\_list), 
event type (event\_type), 
cause (event\_cause), 
process (event\_process), 
outcome (event\_result) \\

Risk \& Impact & 
Risk description (ai\_risk\_description), 
risk type (ai\_risk\_type), 
risk subtype (ai\_risk\_subtype), 
realization status (realized\_or\_potential), 
harm type (harm\_type), 
severity (harm\_severity), 
affected actor type (affected\_actor\_type), 
affected actor subtype (affected\_actor\_subtype), 
risk stage (risk\_stage) \\
\bottomrule
\end{tabularx}
\end{table*}

To support this decision, the LLM performs joint risk identification and structured attribute extraction in a single-pass inference process. 
Once an article is judged to be AI risk-related, the model simultaneously extracts three groups of attributes: AI technology elements, AI risk elements, and event elements. 
These attributes include AI systems, risk descriptions, harm types, affected actors, temporal information, location, core actors, and event outcomes. 
The extracted structure provides the empirical basis for determining whether the article contains a sufficiently complete and grounded AI risk incident. The extracted attributes used in this stage are summarized in Table~\ref{tab:extraction_schema}.

In particular, an article is retained as an event-level AI risk incident only if it contains the minimum event anchors required for reliable downstream alignment, including a grounded core actor, an identifiable incident type, a related AI system, and at least one of the following contextual anchors: temporal information or location information. 
Articles that are AI risk-related but do not satisfy these grounding conditions are categorized as discussion-level reports.

\subsection{Incident Alignment}\label{subsec2_alignment}

\begin{figure}
    \centering
    \includegraphics[width=1.0\linewidth]{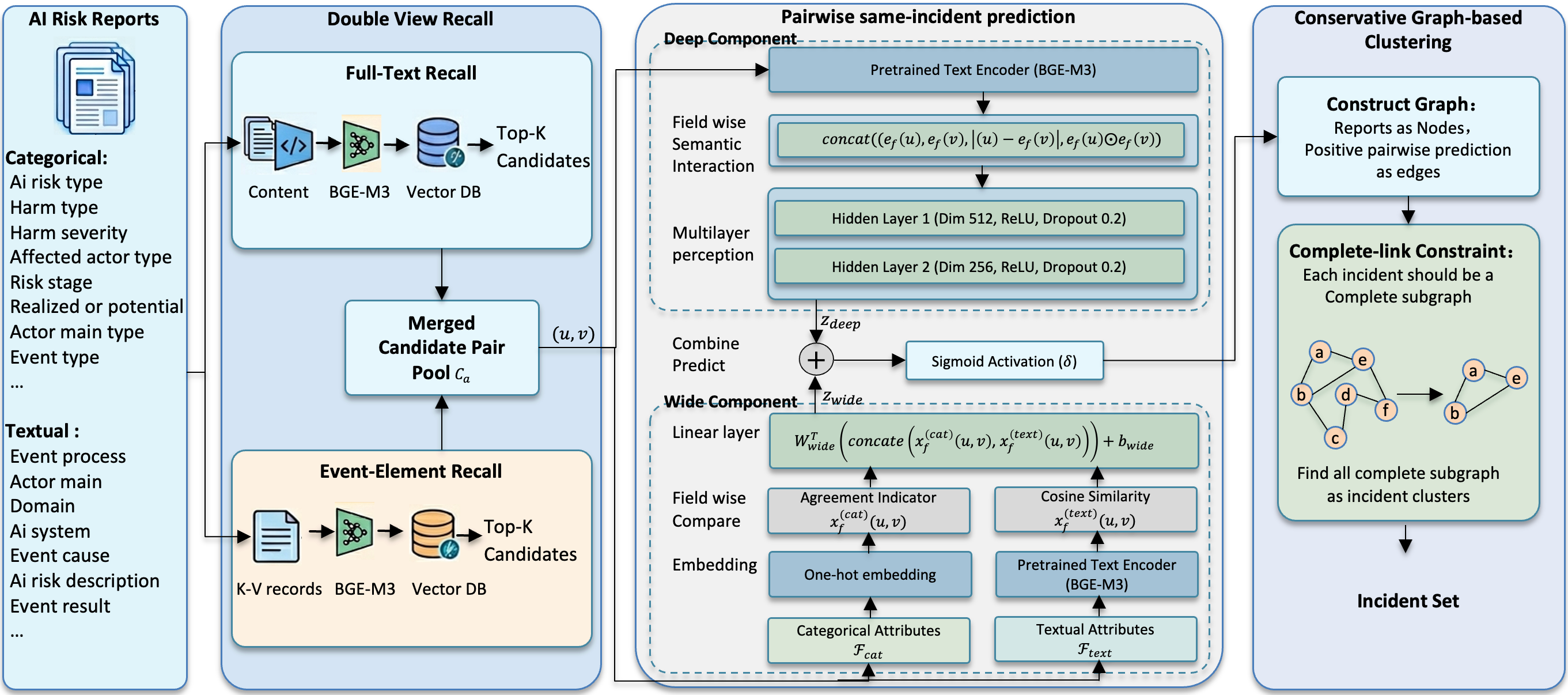}
    \caption{The framework of incident alignment.}
    \label{fig:incident-alignment}
\end{figure}

Incident alignment is performed on the event-level AI risk incident articles identified in the previous stage, denoted as \(\mathcal{A}_{incident}\). 
Each article in \(\mathcal{A}_{incident}\) is associated with structured event elements extracted during identification, including temporal and location information, involved actors, related AI systems, event type, cause, process, and result. 
These grounded attributes provide the basis for cross-document incident matching.

To aggregate reports that refer to the same real-world incident, we adopt a three-stage workflow (shown in Fig.~\ref{fig:incident-alignment}) consisting of dual-view candidate recall, pairwise same-incident prediction, and conservative graph-based clustering.

\subsubsection{Dual-View Candidate Recall}\label{subsubsec2_candidate}

Given the scale of \(\mathcal{A}_{incident}\), exhaustive pairwise comparison is computationally impractical. 
We therefore first perform coarse-grained candidate recall and apply the pairwise classifier only to a small candidate pool for each query report.

For each article \(a \in \mathcal{A}_{incident}\), we construct two complementary retrieval views.

\paragraph{(1) Full-text recall.}
We encode the full news content using BGE-M3 and store all report embeddings in a vector database. 
For each query report \(a\), we retrieve the top \(K\) reports with the highest cosine similarity in the full-text embedding space. 
This view preserves narrative details and broader contextual semantics.

\paragraph{(2) Event-element recall.}
We further construct a structured event representation by concatenating the extracted incident attributes (shown in Table ~\ref{tab:extraction_schema}) into a normalized key--value style text, including event actors, AI systems, event type, event cause, event process, event result, and AI risk attributes. 
This structured representation is also encoded using BGE-M3, and for each query report \(a\), we retrieve the top \(K\) most similar reports in this event-element embedding space. 
Compared with raw news text, this view suppresses background details unrelated to the incident and highlights the core event structure.

Formally, let \(\mathbf{v}_{a,\text{text}}\) and \(\mathbf{v}_{a,\text{event}}\) denote the full-text embedding and event-element embedding of report \(a\), respectively. 
For a query report \(a\) and candidate report \(b\), we compute cosine similarity:
\[
s_{\text{text}}(a, b) =
\frac{\mathbf{v}_{a,\text{text}} \cdot \mathbf{v}_{b,\text{text}}}
{\|\mathbf{v}_{a,\text{text}}\| \|\mathbf{v}_{b,\text{text}}\|}, \quad
s_{\text{event}}(a, b) =
\frac{\mathbf{v}_{a,\text{event}} \cdot \mathbf{v}_{b,\text{event}}}
{\|\mathbf{v}_{a,\text{event}}\| \|\mathbf{v}_{b,\text{event}}\|}.
\]

The final candidate pool is formed by merging the top \(K\) results from both channels:
\[
\mathcal{C}_a =
\{b \in \mathcal{A}_{incident}\mid \mathrm{rank}_{s_{\text{text}}}(a,b) \le K\}
\cup
\{b \in \mathcal{A}_{incident}\mid \mathrm{rank}_{s_{\text{event}}}(a,b) \le K\}.
\]

We set \(K=150\) for each view. 
This choice is conservative enough to maximize recall of same-incident reports while keeping the downstream pairwise classification cost manageable. 
In the AIID--AIAAIC training data, most incidents contain only a limited number of associated reports, and even relatively large incidents rarely exceed 40--50 reports.

\subsubsection{Pairwise Same-Incident Prediction via a Simplified DeepWide Model}

Given a candidate pair \((u, v)\), we formulate same-incident prediction as a binary classification problem and estimate \(P(y=1 \mid u, v)\), where \(y=1\) indicates that the two reports describe the same real-world incident. 
We adopt a simplified DeepWide architecture that combines explicit matching signals derived from structured features with latent semantic interactions captured by dense representations.

\paragraph{Wide Component: Explicit Matching Features}

The wide branch encodes interpretable, low-dimensional features that directly measure cross-report consistency.

Let \(\mathcal{F}_{cat}\) denote the set of categorical attributes:
\[
\mathcal{F}_{cat} =
\left\{
\begin{aligned}
&\textit{ai\_risk\_type},\ \textit{harm\_type},\ \textit{harm\_severity},\ \textit{affected\_actor\_type},\\
&\textit{risk\_stage},\ \textit{realized\_or\_potential},\ \textit{actor\_main\_type}
\end{aligned}
\right\}.
\]

For each \(f \in \mathcal{F}_{cat}\), we define a binary agreement indicator:
\[
x_f^{(cat)}(u,v)=
\begin{cases}
1, & \text{if } u_f=v_f,\\
0, & \text{otherwise}.
\end{cases}
\]

Let \(\mathcal{F}_{text}\) denote the set of textual attributes:
\[
\mathcal{F}_{text} =
\left\{
\begin{aligned}
&\textit{event\_process},\ \textit{actor\_main},\ \textit{domain},\ \textit{ai\_system},\ \textit{event\_type},\\
&\textit{event\_cause},\ \textit{ai\_risk\_description},\ \textit{event\_result},\\
&\textit{ai\_risk\_subtype},\ \textit{affected\_actor\_subtype}
\end{aligned}
\right\}.
\]

For each \(f \in \mathcal{F}_{text}\), we encode the corresponding field using a pretrained text encoder, producing
\[
\mathbf{e}_f(u),\ \mathbf{e}_f(v) \in \mathbb{R}^d,
\]
and compute the cosine similarity:
\[
x_f^{(text)}(u,v)=
\frac{\mathbf{e}_f(u)\cdot \mathbf{e}_f(v)}
{\|\mathbf{e}_f(u)\|\,\|\mathbf{e}_f(v)\|}.
\]

The complete wide feature vector is
\[
\mathbf{x}_{wide}(u,v)=
\mathrm{concat}\Big(
\{x_f^{(cat)}(u,v)\}_{f\in\mathcal{F}_{cat}},
\{x_f^{(text)}(u,v)\}_{f\in\mathcal{F}_{text}}
\Big),
\]
and the wide branch output is
\[
z_{wide}=\mathbf{w}_{wide}^{\top}\mathbf{x}_{wide}(u,v)+b_{wide}.
\]

\paragraph{Deep Component: Field-wise Semantic Interaction}

The deep branch models higher-order semantic relations through fine-grained field-wise interactions. 
For each field \(f \in \mathcal{F}_{text}\), we obtain embeddings
\[
\mathbf{e}_f(u)=\mathrm{Enc}(u_f), \qquad 
\mathbf{e}_f(v)=\mathrm{Enc}(v_f),
\]
and construct the deep input as
\[
\mathbf{x}_{deep}(u,v)=
\mathrm{concat}\Big(
\big\{
\mathbf{e}_f(u),\;
\mathbf{e}_f(v),\;
|\mathbf{e}_f(u)-\mathbf{e}_f(v)|,\;
\mathbf{e}_f(u)\odot \mathbf{e}_f(v)
\big\}_{f \in \mathcal{F}_{text}}
\Big).
\]

This representation is then passed through a multilayer perceptron:
\[
z_{deep}=\mathrm{MLP}(\mathbf{x}_{deep}(u,v)).
\]

\paragraph{Final Prediction}

The final prediction is obtained by combining the wide and deep outputs:
\[
P(y=1\mid u,v)=\sigma(z_{wide}+z_{deep}).
\]

\paragraph{Implementation and Training.}

The wide branch is implemented as a single linear layer over explicit matching features, without additional hidden layers. 
The deep branch uses a two-layer multilayer perceptron with hidden dimensions 512 and 256, respectively; each hidden layer is followed by a ReLU activation and dropout with rate 0.2.

We use BGE-M3 as the pretrained encoder for field-level textual similarity in the wide branch and for field-wise embeddings in the deep branch. 
The model is trained using binary cross-entropy loss:
\[
\mathcal{L}=-y\log \hat{y}-(1-y)\log(1-\hat{y}),
\]
where \(\hat{y}=P(y=1\mid u,v)\). 
We use AdamW as the optimizer with a learning rate of \(2\times10^{-4}\) and a batch size of 64, and apply early stopping based on validation F1 score.

To train the pairwise predictor, we convert the AIID--AIAAIC event--news dataset (over 2,000 incidents and 6,700+ reports) into a report-pair dataset. 
Positive pairs are report pairs belonging to the same incident. 
Negative pairs are constructed through hard negative sampling: for each query report, we first recall its top-\(K\) candidate reports using the same dual-view recall strategy, and then select high-ranking but incorrect candidates as negatives. 
The final training set uses a positive-to-negative ratio of 1:3.

\subsubsection{Conservative Graph-Based Clustering}\label{subsubsec2_cluster}

Pairwise same-incident predictions are converted into incident clusters using a conservative graph-based strategy. 
Each report is treated as a node, and each positive pairwise prediction is interpreted as an edge.

To ensure high intra-cluster consistency, we impose a complete-link constraint.

\paragraph{Complete-link constraint.}
When adding a report \(b\) to an existing cluster \(C\), we require that \(b\) be positively linked to every report already in \(C\). 
In other words, the augmented cluster must remain a complete graph, ensuring that all reports within a cluster are strongly mutually supported as belonging to the same incident.

We adopt this conservative strategy rather than transitive connectivity. 
Under a transitive merging rule, if \(a\) links to \(b\) and \(b\) links to \(c\), then \(a\) and \(c\) would be merged into the same cluster by default. 
In practice, this leads to severe error propagation and unrealistically large clusters. 
In preliminary experiments, such a permissive strategy produced a giant cluster containing more than 260,000 reports. 
By contrast, the complete-link constraint effectively reduces cross-incident merging and yields a more reliable incident-centered structure.

\subsection{Multi-dimensional Risk Classification}\label{subsec2_taxonomy}

To transform aligned incidents into structured and analyzable representations, we introduce a unified multi-dimensional risk classification framework. 
This framework integrates regulatory and academic perspectives, primarily the EU \textit{Artificial Intelligence Act (AI Act)} and the MIT Media Lab taxonomy used in the \textit{AI Incident Database (AIID)}.

\paragraph{Taxonomy design.}
Each aligned incident is annotated along three groups of dimensions:
\begin{itemize}
  \item \textbf{Risk Level:} one of four levels defined by the EU AI Act---\textit{Unacceptable}, \textit{High}, \textit{Limited}, and \textit{Minimal} Risk;
  \item \textbf{Domain Tags:} based on the MIT taxonomy, including 7 primary domains and 25 subdomains (multi-label);
  \item \textbf{Causal Tags:} defined as a triplet of \textit{Entity} (AI system or human), \textit{Intent} (intentional or unintentional), and \textit{Timing} (pre- or post-deployment), yielding 8 causal types.
\end{itemize}
Each incident is assigned one risk level, one causal tag, and one or more domain tags, forming a consistent multi-dimensional representation of AI risk.

\paragraph{Annotation framework.}
We construct a benchmark of 2,285 aligned incidents using a hybrid annotation strategy that combines LLM-based pre-annotation with expert calibration. 
Each incident is consistently represented by a fixed set of representative reports (five news summaries), which are aggregated as the input context. 
This representation is used uniformly across training, evaluation, and full-dataset inference to ensure consistency between model learning and deployment.
A large language model first generates preliminary labels following the predefined taxonomy, and the outputs are subsequently reviewed and refined by domain experts based on detailed annotation guidelines. 
Each instance is independently annotated by two experts, with disagreements resolved by a third annotator, resulting in substantial inter-annotator agreement (Cohen's \(\kappa = 0.74\)).

\paragraph{Modeling and evaluation.}
Based on the curated annotations, we formulate incident classification as a combination of single-label (risk level and causal tag) and multi-label (domain tags) prediction tasks. 
We adopt supervised instruction tuning for large language models, where each instance follows an instruction--input--output format: the instruction specifies the taxonomy and labeling rules, the input consists of aggregated incident evidence, and the output is a structured set of classification labels. 

To enable efficient adaptation of large models, we employ parameter-efficient fine-tuning using Low-Rank Adaptation (LoRA).

We evaluate both base and fine-tuned models using standard metrics. 
For single-label tasks, we report accuracy and F1 score; for multi-label tasks, we report Hamming loss and micro/macro-averaged metrics. 
Experimental results show that fine-tuned models consistently outperform base models and prompt-based approaches, demonstrating the effectiveness of the proposed taxonomy and classification framework.

\section{Data Records}\label{sec3}

\subsection{Dataset Composition}\label{subsec3_composition}

Table~\ref{tab:dataset} summarizes the current composition of the RiskNet dataset across source collections. In total, the pipeline identified 777.1K AI risk-related reports, of which 264.7K were categorized as event-level reports and used for downstream incident alignment.

\begin{table*}[t]
\centering
\small
\renewcommand{\arraystretch}{1.1}
\caption{Overview of the data sources and the RiskNet dataset}
\label{tab:dataset}
\begin{tabular*}{\textwidth}{@{\extracolsep{\fill}}lcccccc@{}}
\toprule
\textbf{Dataset} & \textbf{Time Range} & \textbf{Lang.} & \textbf{Raw} & \textbf{AI-rel.} & \textbf{AI Risk} & \textbf{Event} \\
\midrule
AIID              & 2016--2025 & Mixed   & 5.0K  & --    & 4.5K   & 3.4K \\
AIAAIC            & 2021--2025 & Mixed   & 4.5K  & --    & 4.1K   & 2.7K \\
\makecell[l]{CommonCrawl News} 
                  & 2022--2025 & English & 674M  & 2.30M & 390K   & 80.2K \\
OpenNewsArchive   & 2023       & Chinese & 8.8M  & 65.6K & 7.9K   & 1.4K \\
News Websites     & 2024       & Chinese & --    & --    & 800    & -- \\
Hot-list Rankings & 2021--2025 & Chinese & 340K  & 4.48K & 790    & -- \\
China News        & 2023--2025 & Chinese & --    & 2.98M & 369K   & 177K \\
\midrule
\textbf{Overall}  & 2016--2025 & Mixed   & \textbf{683M} & \textbf{5.35M} & \textbf{777.1K} & \textbf{264.7K} \\
\bottomrule
\end{tabular*}
\end{table*}

The source composition highlights the complementary roles of public incident repositories, large-scale open news archives, online trend signals, and licensed news collections.

\subsection{Incident-level Alignment Records}\label{subsec3_alignment_records}

Applying the alignment workflow to the full event-level corpus yielded \textbf{54,386} incident clusters from \textbf{264,776} event-level reports. As shown in Table~\ref{tab:full_data_stats}, the resulting incident--news structure has an average cluster size of \textbf{4.87} reports, with substantial variability across incidents.

\begin{table}[ht]
\centering
\small
\setlength{\tabcolsep}{8pt}
\renewcommand{\arraystretch}{1.15}
\caption{Descriptive statistics of the aligned incident--reports structure}
\label{tab:full_data_stats}
\begin{tabular}{@{}p{4.2cm}r@{}}
\toprule
\textbf{Statistic} & \textbf{Value} \\
\midrule
Total event-level reports & 264,776 \\
Aligned incident clusters (\(K\)) & 54,386 \\
Singleton clusters (\(N=1\)) & 16,197 \\
\midrule
Maximum cluster size & 151 \\
Average cluster size (\(\mu\)) & 4.87 \\
Standard deviation (\(\sigma\)) & 8.99 \\
\bottomrule
\end{tabular}
\end{table}

\begin{figure}
    \centering
    \includegraphics[width=1.0\linewidth]{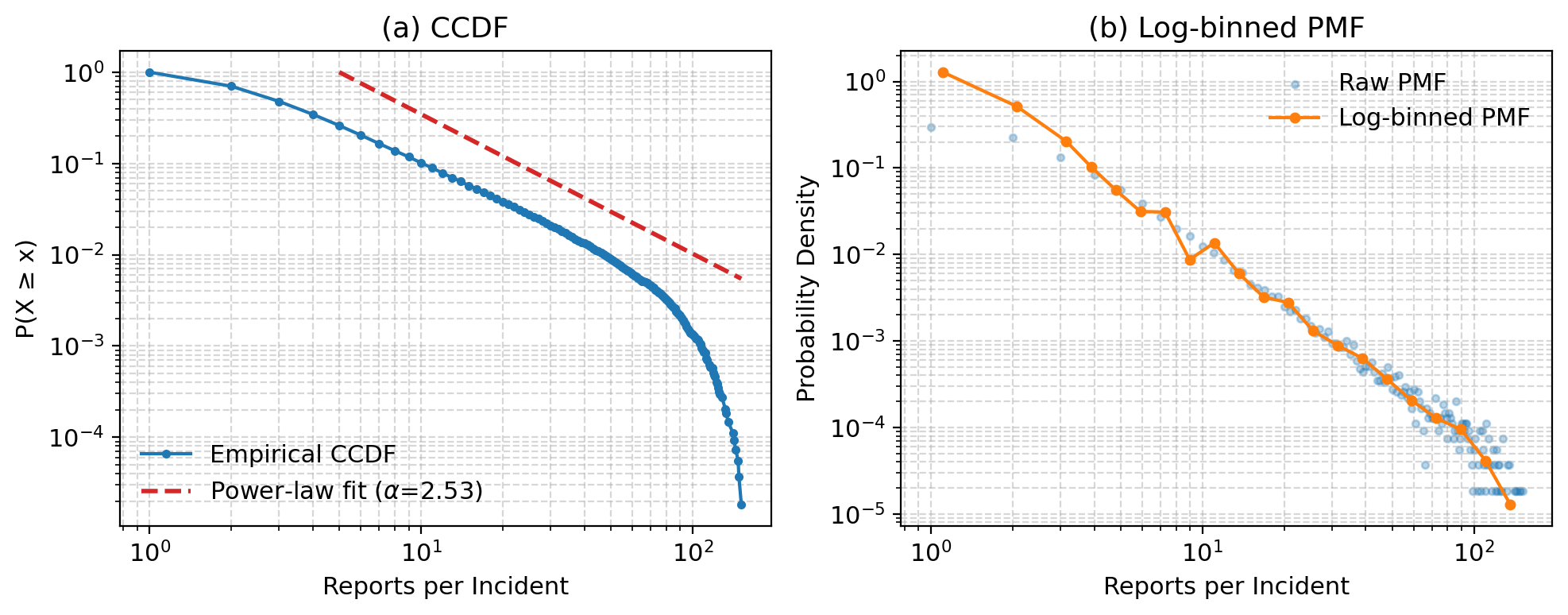}
    \caption{Distribution of incident cluster sizes: (a) CCDF with power-law fit ($\alpha \approx 2.53$); (b) log-binned PMF.}
    \label{fig:full_event_data}
\end{figure}

The cluster-size distribution exhibits a clear heavy-tailed pattern, as illustrated in Figure~\ref{fig:full_event_data}. 
Specifically, 29.8\% (16,197) of the identified incidents are singletons, while the majority (66.7\%) form small-to-medium clusters containing 2 to 20 reports. 
In contrast, a small number of incidents are associated with substantially larger numbers of reports, corresponding to highly publicized AI risk events.

Figure~\ref{fig:full_event_data}(a) shows the complementary cumulative distribution function (CCDF) on log--log scales, where the tail approximately follows a straight line, indicating power-law behavior. 
Fitting the tail yields a scaling exponent of $\alpha \simeq 2.53$, suggesting a truncated heavy-tailed distribution. 
Figure~\ref{fig:full_event_data}(b) further presents the log-binned probability mass function (PMF), which exhibits an approximately linear decay in the mid-range and increased variance in the tail region, consistent with finite-size effects and data sparsity.

Together, these observations indicate that while most incidents receive limited coverage, a small fraction attract disproportionately large attention, reflecting the heterogeneous visibility of AI risk events across media sources.

\subsection{Classification Records}\label{subsec3_classification_records}

\begin{figure}[H]
    \centering
    \includegraphics[width=1\textwidth]{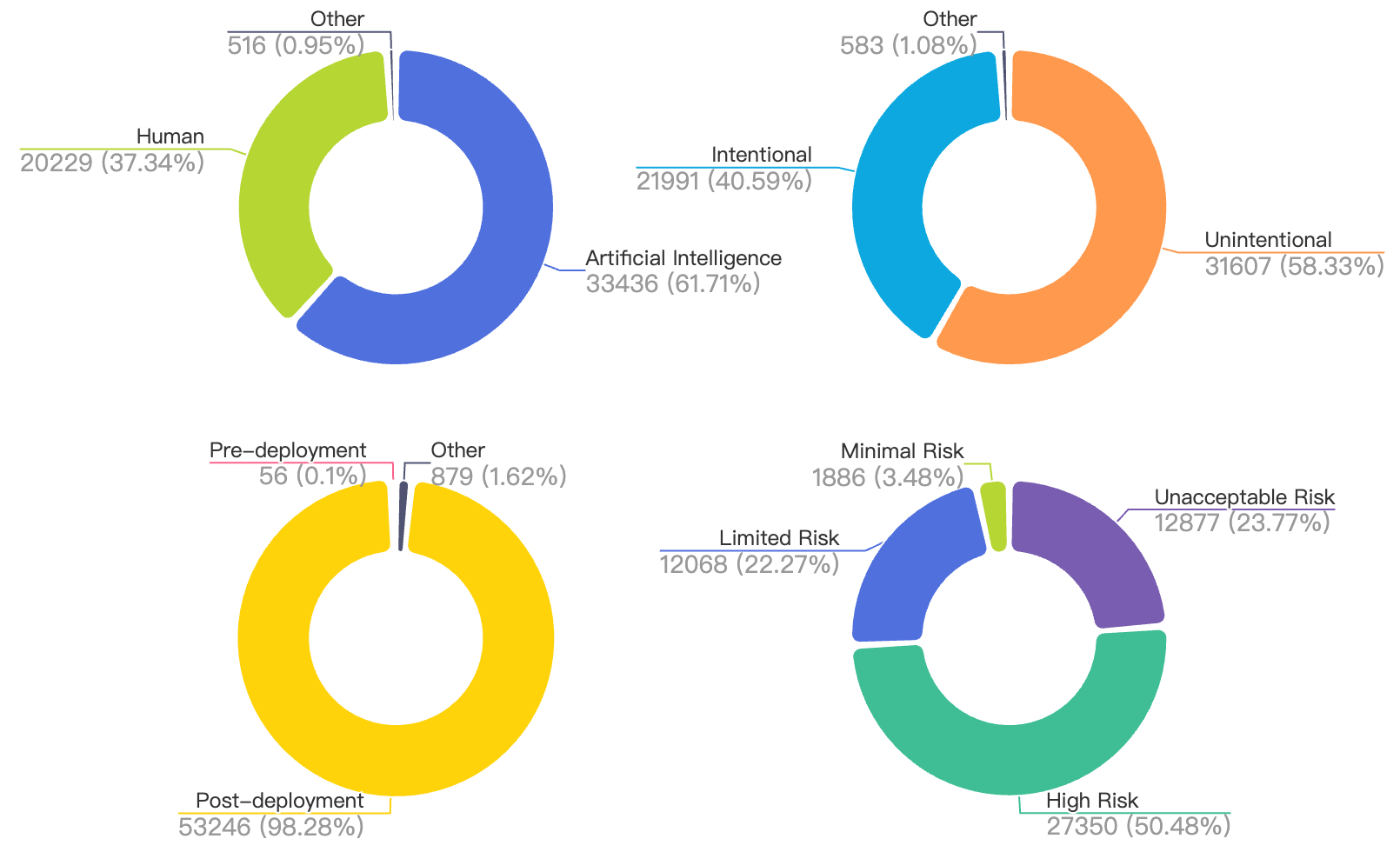}
    \caption{The distributions of risk incidents on  intention, timing, entity, and level.}
    \label{fig:unilabel-clas}
\end{figure}

\begin{figure}[H]
    \centering
    \includegraphics[width=1\textwidth]{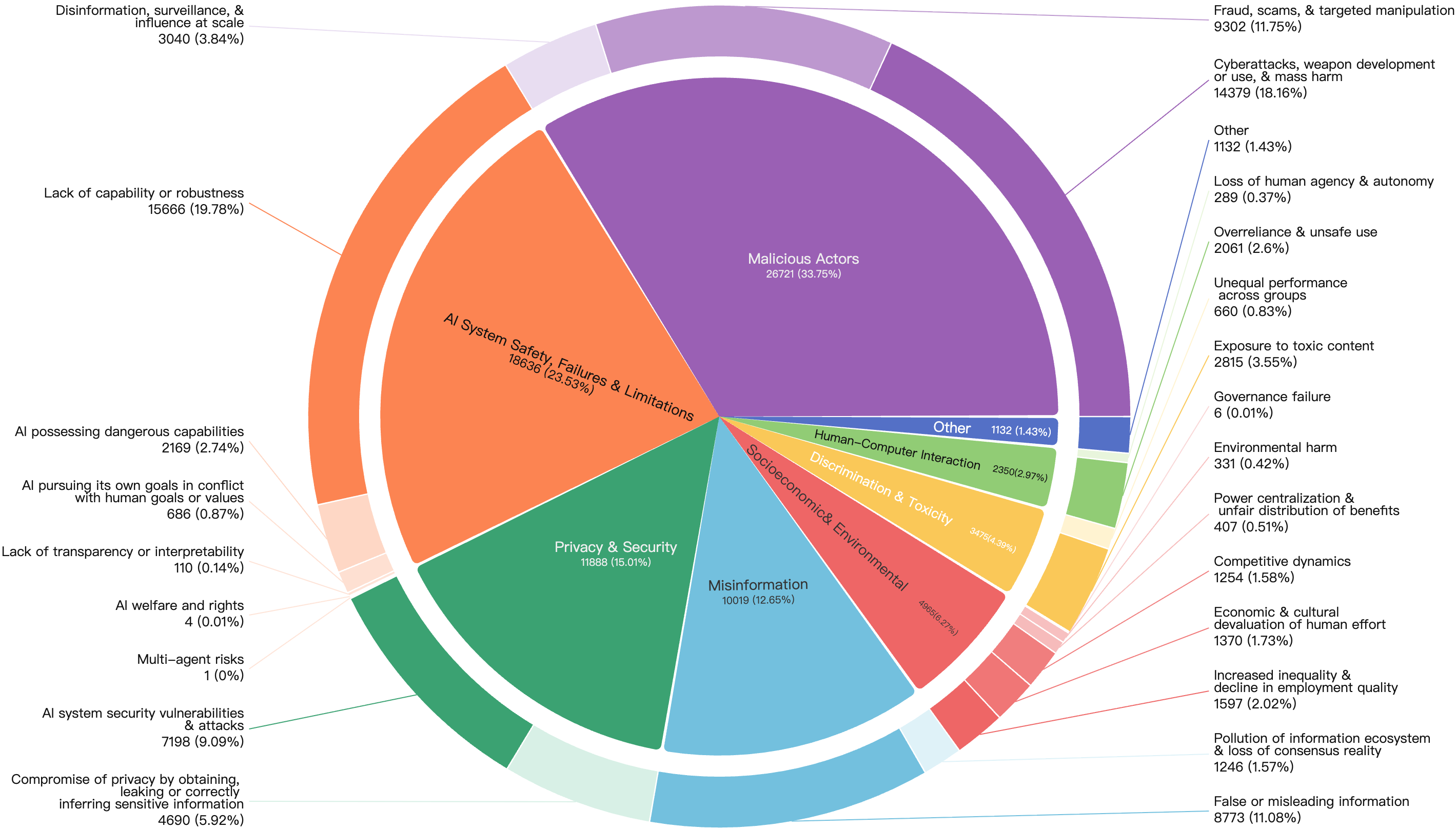}
    \caption{The distributions of risk incidents on domain labels.}
    \label{fig:multilabel-clas}
\end{figure}

We applied the incident classification pipeline to the aligned RiskNet incidents and summarized their label distributions across multiple dimensions. Figures~\ref{fig:unilabel-clas} and \ref{fig:multilabel-clas} show the resulting distributions over causal tags, timing, entity, risk level, and domain labels.

The most frequent subdomains are summarized in Table~\ref{tab:top5domain}. These descriptive statistics provide an overview of the empirical composition of the dataset and illustrate the diversity of AI risk incidents captured in RiskNet.

\begin{table}[!htbp]
\centering
\small
\caption{Most frequent risk subdomains in the classified incident dataset}
\label{tab:top5domain}
\begin{tabular}{lc}
\toprule
\textbf{Subdomain} & \textbf{Proportion} \\
\midrule
Lack of capability or robustness & 19.78\% \\
Cyberattacks, weapon development or use \& mass harm  & 18.16\% \\
Fraud, scams \& targeted manipulation & 11.75\% \\
False or misleading information & 11.08\% \\
AI system security vulnerabilities \& attack & 9.09\% \\
Compromise of privacy by obtaining, leaking or inferring & 5.92\% \\
\bottomrule
\end{tabular}
\end{table}

\subsection{Data Organization and Key Fields}\label{subsec3_dataset_description}

RiskNet is organized around three types of released information: 
(1) source news records with basic metadata, 
(2) structured fields extracted during AI risk identification, and 
(3) incident-level linkage and classification labels derived from incident alignment and downstream annotation.

\paragraph{(1) Source news records and metadata.}
Each source record preserves the original article-level information collected from the upstream repositories. 
The core metadata fields include the source name, publication time, title, article content, and original URL when available. 
These fields provide the raw textual evidence and provenance information for subsequent identification, alignment, and analysis. 
Each record is also assigned a stable news identifier for downstream linkage.

\paragraph{(2) Extracted structured fields.}
During the AI risk identification stage, each relevant record is further annotated with structured fields describing the AI system, event context, and risk impact. 
These extracted attributes are used both to distinguish event-level reports from discussion-level reports and to support candidate recall and pairwise matching in incident alignment. 
The full extraction schema has been introduced in Table~\ref{tab:extraction_schema}, including three dimensions: \textit{AI System \& Application}, \textit{Event Context}, and \textit{Risk \& Impact}. 
To avoid redundancy, we do not repeat the full field list here.

\paragraph{(3) Incident linkage and classification labels.}
After incident alignment, each event-level news record is linked to an aligned incident identifier, which represents the real-world incident cluster to which the report belongs. 
This mapping establishes the connection between article-level evidence and incident-level analysis units. 
In addition, each aligned incident is associated with multi-dimensional classification labels, including entity, intent, time, risk level, domain, and subdomain. 
These labels support downstream tasks such as incident analysis, benchmark construction, and structured risk assessment.

Overall, the released data therefore connect raw article evidence, extracted incident-related attributes, and aligned incident-level labels within a unified schema.

\section{Technical Validation}\label{sec4}

\subsection{Validation of AI Risk News Identification}\label{subsec4_identification_validation}

We assessed the quality of the identification workflow at two stages.

\begin{itemize}
    \item \textbf{Stage 1: AI Relevance Filtering.}  
    We used the union of AIID and AIAAIC as a reference set to evaluate whether the keyword-based filtering stage could successfully recall known AI-related incident reports. Among 9,561 reference articles, 9,138 were successfully recalled, corresponding to a recall of 0.9557.

    \item \textbf{Stage 2: AI Risk Incident Identification.}  
    We randomly sampled 2,000 AI-related articles and annotated them manually with three human annotators. Using this benchmark, the LLM-based identification stage achieved a precision of 0.9550 and a recall of 0.9717.
\end{itemize}

The results are summarized in Table~\ref{tab:filtering_combined_results}.
\begin{table}[t]
\centering
\small
\setlength{\tabcolsep}{6pt}
\renewcommand{\arraystretch}{1.2}
\caption{Performance of AI risk identification}
\label{tab:filtering_combined_results}
\begin{tabular}{@{}lccc@{}}
\toprule
\textbf{Stage} & \textbf{Size} & \textbf{Precision} & \textbf{Recall} \\
\midrule
AI relevance filtering & 9,561 & --     & 0.9557 \\
Risk identification    & 2,000 & 0.9550 & 0.9717 \\
\bottomrule
\end{tabular}
\end{table}

\subsection{Validation Benchmark Construction}\label{subsec4_benchmark}

To validate the quality of the event-level distinction and incident alignment workflow, we constructed two benchmark subsets: an event classification subset and an incident alignment subset.

\subsubsection{Event Classification Benchmark}\label{subsubsec4_event_subset}

This subset was constructed to validate the distinction between event-level reports, discussion-level reports, and irrelevant AI news. We randomly sampled 2,000 reports from a large-scale corpus of AI-related news. Two annotators independently labeled each report, and disagreements were adjudicated by a third annotator.

The three labels are defined as follows:
\begin{itemize}
    \item \textbf{AIrisk\_relevant\_event}: a report describing a concrete and localized AI risk incident with sufficient event grounding;
    \item \textbf{AIrisk\_relevant\_discussion}: a report discussing AI-related risks without sufficient incident grounding for alignment;
    \item \textbf{AIrisk\_Irrelevant}: AI-related news that does not concern risk.
\end{itemize}

The class distribution is shown in Table~\ref{tab:event_subset_distribution}.

\begin{table}[t]
\centering
\small
\setlength{\tabcolsep}{8pt}
\renewcommand{\arraystretch}{1.3}
\caption{AI risk report classification}
\label{tab:event_subset_distribution}
\begin{tabular}{@{}lrr@{}}
\toprule
\textbf{Class} & \textbf{Count} & \textbf{Share (\%)} \\
\midrule
Irrelevant AI content    & 1,400 & 70.00 \\
Event-level AI risk      & 193   & 9.65 \\
Discussion-level AI risk & 407   & 20.35 \\
\bottomrule
\end{tabular}
\end{table}

\subsubsection{Incident Alignment Benchmark}\label{subsubsec4_alignment_subset}

This subset was constructed to validate the quality of incident alignment against reference event--news associations available in AIID and AIAAIC. We first merged overlapping entries from the two repositories through embedding-based deduplication followed by manual review. We then applied the event-level screening criteria to the merged pool of 9,561 raw news reports, retaining 6,124 event-level reports linked to 1,752 reference incidents.

For pairwise evaluation, positive pairs were defined as two reports assigned to the same reference incident, while negative pairs were sampled from top-ranked retrieval candidates that belonged to different incidents. 


\subsection{Validation of Event-level Report Classification}\label{subsec4_event_validation}

Table~\ref{tab:cat_results} summarizes the performance of the event-level report classification step. The method achieved an overall accuracy of 0.9300. Precision was highest for the \textit{AIrisk\_Irrelevant} category (0.9874), indicating strong ability to filter non-risk AI news. For the \textit{AIrisk\_relevant\_event} category, the method achieved a precision of 0.9018 and a recall of 0.7617, reflecting a conservative operating point that prioritizes the reliability of event-level records retained for downstream incident alignment.

\begin{table}[t]
\centering
\small
\setlength{\tabcolsep}{6pt}
\renewcommand{\arraystretch}{1.2}
\caption{Performance of the event classification method}
\label{tab:cat_results}
\begin{tabular}{@{}lcccc@{}}
\toprule
\textbf{Class} & \textbf{Precision} & \textbf{Recall} & \textbf{F1} & \textbf{Support} \\
\midrule
Irrelevant AI content    & 0.9874 & 0.9479 & 0.9672 & 1,400 \\
Event-level AI risk      & 0.9018 & 0.7617 & 0.8258 & 193 \\
Discussion-level AI risk & 0.7830 & 0.9484 & 0.8578 & 407 \\
\midrule
\textbf{Accuracy} & \multicolumn{4}{c}{\textbf{0.9300}} \\
\bottomrule
\end{tabular}
\end{table}

\subsection{Validation of Incident Alignment}\label{subsec4_alignment_validation}

We evaluated incident alignment at both the pairwise and cluster levels. Pairwise metrics were used to assess the reliability of same-incident edge prediction, while cluster-level evaluation was used as the primary indicator of incident reconstruction quality.

Let \(\mathcal{G} = \{G_i\}_{i=1}^{K_{\text{gold}}}\) denote the gold incident clusters and \(\mathcal{P} = \{P_j\}_{j=1}^{K_{\text{pred}}}\) the predicted clusters. For each potential correspondence \((G_i, P_j)\), overlap was measured using set-based F1:
\[
S_{ij} = \frac{2|G_i \cap P_j|}{|G_i| + |P_j|}.
\]
To compare predicted and gold clusters under label permutation, we used the Hungarian algorithm to compute an optimal one-to-one assignment. The primary metric was Hungarian Macro-F1:
\[
\text{Hungarian Macro-F1} = \frac{1}{K_{\text{gold}}} \sum_{i=1}^{K_{\text{gold}}} S_{i,\pi(i)}.
\]

Under the selected operating configuration, the alignment workflow achieved a Hungarian Macro-F1 of 0.895 on the test set (Table~\ref{tab:event_metrics}). Pairwise diagnostic metrics are reported in Table~\ref{tab:pair_metrics}.

\begin{table}[ht]
\centering
\small
\setlength{\tabcolsep}{8pt}
\renewcommand{\arraystretch}{1.35}
\caption{Pairwise classification performance on the test set}
\label{tab:pair_metrics}
\begin{tabular}{l c}
\toprule
\textbf{Metric} & \textbf{Value} \\
\midrule
Edge rule & mutual \\
Merging strategy & complete\_link \\
Threshold \(\tau\) & 0.20 \\
\midrule
Positive precision (\(\mathrm{P}_{+}\)) & 0.506 \\
Positive recall (\(\mathrm{R}_{+}\)) & 0.822 \\
Positive F1 (\(\mathrm{F1}_{+}\)) & 0.626 \\
Macro-F1 & 0.803 \\
Accuracy & 0.963 \\
\midrule
Number of evaluated pairs & 244{,}489 \\
Number of predicted positive pairs & 14{,}959 \\
\bottomrule
\end{tabular}
\end{table}

\begin{table}[ht]
\centering
\small
\setlength{\tabcolsep}{8pt}
\renewcommand{\arraystretch}{1.35}
\caption{Incident-level alignment performance on the test set}
\label{tab:event_metrics}
\begin{tabular}{l c}
\toprule
\textbf{Metric} & \textbf{Value} \\
\midrule
Hungarian macro-F1 & 0.895 \\
\midrule
Number of gold incidents (\(K_{\mathrm{gold}}\)) & 351 \\
Number of predicted clusters (\(K_{\mathrm{pred}}\)) & 476 \\
Cluster-count ratio (\(K_{\mathrm{pred}}/K_{\mathrm{gold}}\)) & 1.36 \\
\bottomrule
\end{tabular}
\end{table}

\subsection{Validation of Classification Labels}\label{subsec4_classification_validation}

To assess the quality of the classification benchmark and provide reference baselines for downstream reuse, we evaluated two label inference strategies: prompt-based inference and fine-tuned LLMs. Incident inputs consisted of the top five representative news reports associated with each aligned incident. The benchmark was split into training and test subsets using an 80/20 partition.

Single-label dimensions (causal tag and risk level) were evaluated using Accuracy and F1-score. Multi-label dimensions (domain and subdomain) were evaluated using Hamming Loss, subset Accuracy, and Micro/Macro-F1.

\setlength{\tabcolsep}{4pt}
\begin{table}[t]
\centering
\small
\caption{Performance on causal and risk level classification}
\label{tab:causal_risk_results}
\begin{tabularx}{\linewidth}{@{}lYYYYYYYY@{}}
\toprule
\textbf{Model} & \multicolumn{2}{c}{\textbf{Entity}} & \multicolumn{2}{c}{\textbf{Intent}} & \multicolumn{2}{c}{\textbf{Timing}} & \multicolumn{2}{c}{\textbf{Risk Level}} \\
\cmidrule(lr){2-3} \cmidrule(lr){4-5} \cmidrule(lr){6-7} \cmidrule(lr){8-9}
 & Acc & F1 & Acc & F1 & Acc & F1 & Acc & F1 \\
\midrule
qwen3-32B-sft    & \textbf{0.8216} & \textbf{0.5418} & \textbf{0.7907} & \textbf{0.5816} & 0.9251 & \textbf{0.5756} & 0.5144 & 0.3871 \\
qwen3-14B-sft    & 0.8031 & 0.5261 & 0.7832 & 0.5369 & \textbf{0.9358} & 0.3441 & \textbf{0.5402} & \textbf{0.4019} \\
kimi-k2-prompt   & 0.6871 & 0.3772 & 0.7505 & 0.5133 & 0.9234 & 0.3215 & 0.4956 & 0.3692 \\
qwen3-32B-prompt & 0.6538 & 0.3227 & 0.7380 & 0.5110 & 0.9134 & 0.3448 & 0.5298 & 0.3501 \\
qwen3-14B-prompt & 0.6450 & 0.2785 & 0.7100 & 0.5224 & 0.9258 & 0.3420 & 0.5117 & 0.3869 \\
\bottomrule
\end{tabularx}
\end{table}

\begin{table}[t]
\centering
\small
\caption{Performance on domain and subdomain classification}
\label{tab:domain_subdomain_results}

\begin{tabularx}{\linewidth}{@{}lYYYY YYYY@{}}
\toprule
\textbf{Model} & \multicolumn{4}{c}{\textbf{Domain}} & \multicolumn{4}{c}{\textbf{Subdomain}} \\
\cmidrule(lr){2-5} \cmidrule(lr){6-9}
 & Hamming Loss & Acc & Micro-F1 & Macro-F1 & Hamming Loss & Acc & Micro-F1 & Macro-F1 \\
\midrule
qwen3-32B-sft & \textbf{0.1533} & \textbf{0.8467} & 0.6572 & 0.5865 & 0.0680 & 0.9320 & 0.5595 & 0.3401 \\
qwen3-14B-sft & 0.1556 & 0.8447 & \textbf{0.6671} & \textbf{0.6091} & \textbf{0.0657} & \textbf{0.9344} & \textbf{0.5835} & \textbf{0.3669} \\
kimi-k2-prompt & 0.2423 & 0.7577 & 0.5990 & 0.5688 & 0.1640 & 0.8360 & 0.1020 & 0.0748 \\
qwen3-32B-prompt & 0.2523 & 0.7477 & 0.5896 & 0.5630 & 0.1704 & 0.8296 & 0.0692 & 0.0485 \\
qwen3-14B-prompt & 0.2580 & 0.7369 & 0.5746 & 0.5471 & 0.1426 & 0.8575 & 0.0822 & 0.0516 \\
\bottomrule
\end{tabularx}
\end{table}

Tables~\ref{tab:causal_risk_results} and \ref{tab:domain_subdomain_results} provide baseline performance on the classification benchmark. Fine-tuned models consistently outperform prompt-based inference across most dimensions, indicating that the released benchmark captures stable and learnable regularities in incident-level labeling.

We used the qwen3-14B-sft to classify all of the incidents.

\section{Usage Notes}\label{sec5}

RiskNet is intended as a reusable resource for studying empirical AI risk incidents across time, languages, sources, and risk dimensions. Because the dataset is built from heterogeneous public and licensed news collections and includes multiple stages of automated processing, several usage considerations should be taken into account.

\subsection{Recommended usage granularity}\label{subsec5_granularity}

Different file types in RiskNet are suitable for different research purposes.

\begin{itemize}
    \item \textbf{Source-level records} are most appropriate for reproducing the identification pipeline and studying source-level reporting characteristics.
    \item \textbf{Event-level records} are suitable for article-level retrieval, event mention analysis, and incident alignment research.
    \item \textbf{Incident-level records} are recommended for most downstream analytical tasks, including temporal trend analysis, cross-domain comparison, and risk distribution studies.
    \item \textbf{Benchmark files} are intended for evaluation and model development on ai risk identification, incident alignment, and multi-dimensional incident classification.
\end{itemize}

For studies focused on real-world incident counts or incident-level distributions, we recommend using the aligned incident-level records rather than raw article counts, because multiple reports may correspond to the same underlying event.

\subsection{Interpretation of aligned incidents}\label{subsec5_alignment_note}

Incident alignment is designed to aggregate reports that refer to the same real-world event, but aligned incidents should still be interpreted as data-driven reconstructions rather than authoritative legal or regulatory case determinations. In particular, some incident clusters may reflect incomplete or evolving media coverage, and a small number of clusters may remain over-split or under-merged despite the conservative alignment strategy.

\subsection{Use of classification labels}\label{subsec5_classification_note}

The classification labels in RiskNet are designed to support structured analysis of AI risk incidents, not to provide definitive normative or legal judgments. Labels such as risk level, domain, and causal category should be interpreted as annotation outputs under the released taxonomy and guidelines.

Two practical implications follow. First, the labels are best suited for aggregate analysis, benchmarking, and comparative research rather than for formal adjudication of individual incidents. Second, incidents assigned to broad categories such as \textit{Other} may indicate emerging or insufficiently captured risk patterns rather than annotation error alone.

\subsection{Temporal and source-related considerations}\label{subsec5_temporal_note}

RiskNet combines records from sources with different publication practices, collection mechanisms, and update frequencies. Publication time may therefore reflect different stages of incident reporting, including early reporting, follow-up coverage, retrospective commentary, or database ingestion time. Users analyzing time trends should consider whether to use source publication time, first-seen incident time, or alignment-based incident time proxies, depending on the research question.

Similarly, differences in source coverage may affect the observed distribution of incidents across domains, countries, and languages. Counts should therefore be interpreted as observations from the available media ecosystem rather than as exhaustive measures of all real-world AI harms.

\subsection{Language and cross-lingual analysis}\label{subsec5_language_note}

Because RiskNet contains multilingual records, cross-lingual analyses should account for possible differences in reporting style, media salience, vocabulary, and source accessibility. For tasks involving semantic retrieval or label modeling across languages, users may benefit from language-specific preprocessing or multilingual normalization.

When cross-country or cross-lingual comparisons are performed, we recommend reporting both incident-level counts and source composition statistics to reduce the risk of over-interpreting source-induced differences.

\subsection{Reuse for benchmarking and modeling}\label{subsec5_benchmark_note}

The benchmark subsets released with RiskNet are suitable for evaluating event-level report classification, incident alignment, and multi-dimensional incident classification. Because benchmark construction partially relies on existing repositories and annotation guidelines, users should preserve the released train--test partitioning or clearly document any re-splitting strategy to maintain comparability.

For machine learning tasks, users should also be aware that semantically similar reports from the same incident can introduce leakage if data are split at the article level rather than the incident level. Incident-wise splitting is therefore recommended for most predictive experiments.

\section*{Data availability}

Code, benchmark datasets, and a subset of AI risk incident data are publicly available at:
\noindent\textbf{Code and datasets:} \url{https://github.com/risk-net}.

The public release includes benchmark annotations for event classification, incident alignment, and multi-dimensional classification, along with sample event-level records to support reproducibility.

The full RiskNet dataset is large in scale, covering millions of source records and hundreds of thousands of aligned reports. Due to data scale and licensing constraints of certain news sources, the complete dataset is not directly released via the repository.

Access to the full aligned dataset is provided through the RiskNet platform (\url{http://www.risknet.group/}) upon request. For restricted-source data, only metadata, extracted attributes, alignment results, and identifiers are released, while original full-text content must be obtained from the respective providers.

\section*{Ethics statement}

This study uses publicly available news data and third-party incident collections for research purposes. No personal sensitive data were intentionally collected.

RiskNet documents reported AI risk incidents derived from media and repository sources. As such, the dataset may contain incomplete, disputed, or evolving accounts of real-world events, including sensitive descriptions of harm or contested responsibility.

Users should exercise caution when interpreting individual records and avoid making unsupported claims about specific persons or organizations. The dataset is intended for research, monitoring, benchmarking, and aggregate analysis, rather than for automated high-stakes decision-making.

For licensed or restricted sources, only metadata and derived annotations are released, without redistributing original copyrighted content. Use of the dataset should comply with applicable legal and ethical standards.

\section*{Usage and licensing}

Public components of RiskNet (code, benchmarks, and sample data) are released under an open-source license specified in the repository.

Access to the full aligned dataset is granted for academic research upon application via the RiskNet platform. 
Users must comply with the licensing terms of the original data sources. 
Users are expected to cite this paper when using RiskNet in academic work.

\section*{Author contributions}

Leihan Zhang conceived the research problem, designed the overall framework and methodology, supervised the study, led the writing and revision of the manuscript, and acquired funding for the project. Weicheng Ye was responsible for data collection and processing, and contributed to the design, implementation, and evaluation of the AI risk identification and incident alignment methods, as well as manuscript preparation. Xianlong Ma designed and implemented the multi-dimensional risk classification method, contributed to system development, and participated in manuscript writing. Haochuan Liu contributed to the design and evaluation of the incident alignment method and participated in system development. Yang Li, Qianyu Zhang, and Jinliang Chen contributed to methodology discussions, system design, and data processing and annotation. Qiang Yan supervised the project, provided research guidance, and contributed to funding acquisition.  All authors read and approved the final manuscript.

\section*{Funding}

This work was supported by the National Natural Science Foundation of China (NSFC) under Grant Nos. 62102044 and 72501034.

\bibliography{bib-Risknet}

\end{document}